\documentclass[graybox]{SNmult}

\usepackage{type1cm}        
\usepackage{makeidx}         
\usepackage{graphicx}        
\usepackage{multicol}        
\usepackage[bottom]{footmisc}

\usepackage{newtxtext}       %
\usepackage[varvw]{newtxmath}       

\makeindex             

\usepackage{url}
\usepackage{tabularx}
\usepackage{booktabs}

\AtBeginDocument{%
  }

\begin{document}
\motto{Our work is a dialogue between the tool and the image, so we would not preconceive an image...We would rather make a tool and dialogue with it — Woody Vasulka, 1971 \cite{gillVideoStateArt1976}}
\title*{Attention of a Kiss: Generative Video Attention Maps as Explainable and Creative Material\thanks{Under review for ``Explainable AI for the Arts'' (N. Bryan-Kinns, Ed.), Springer.}}
\author{Adam Cole\orcidID{0000-0001-9715-314X} and\\ Mick Grierson\orcidID{0000-0002-6981-5414}}
\institute{Adam Cole \at Creative Computing Institute, University of the Arts London, 272 High Holborn, London WC1V 7EY, UK \email{a.cole@arts.ac.uk}
\and Mick Grierson \at Creative Computing Institute, University of the Arts London, 272 High Holborn, London WC1V 7EY, UK \email{m.grierson@arts.ac.uk}}
%
%
\maketitle
\abstract{This paper presents an artistic and technical investigation into the attention mechanisms of video diffusion transformers. Inspired by early video artists who manipulated analog video signals to create new visual aesthetics, this study proposes a method for extracting and visualizing cross-attention maps in generative video models. Built on the open-source Wan model, our tool provides an interpretable window into the temporal and spatial behavior of attention in text-to-video generation. Through exploratory probes and a series of artistic case studies, we examine the potential of attention maps as both analytical tools and raw artistic material. This work contributes to the growing field of \textit{Explainable AI for the Arts} (XAIxArts), inviting artists to reclaim the inner workings of AI as a creative medium.
}

\setcounter{footnote}{0}

\section{Introduction}
Reviewing the history of early video art, a common trend emerges: a drive to produce imagery that disrupted the aesthetic conventions of broadcast television and mainstream cinema \cite{meigh-andrewsHistoryVideoArt2013}. To achieve this, artists like Nam June Paik and the Vasulkas cultivated a deep technical understanding of video systems, enabling them to construct bespoke tools capable of manipulating analog signals in expressive, often radical ways. These interventions yielded profoundly original video works that redefined the medium \cite{meigh-andrewsHistoryVideoArt2013}. With the rise of AI video models, we ask: can a similar strategy be applied today — one that harnesses technical insight to subvert and expand the generative possibilities of these new systems?

\begin{figure}[t]
  \centering
  \includegraphics[width=\linewidth]{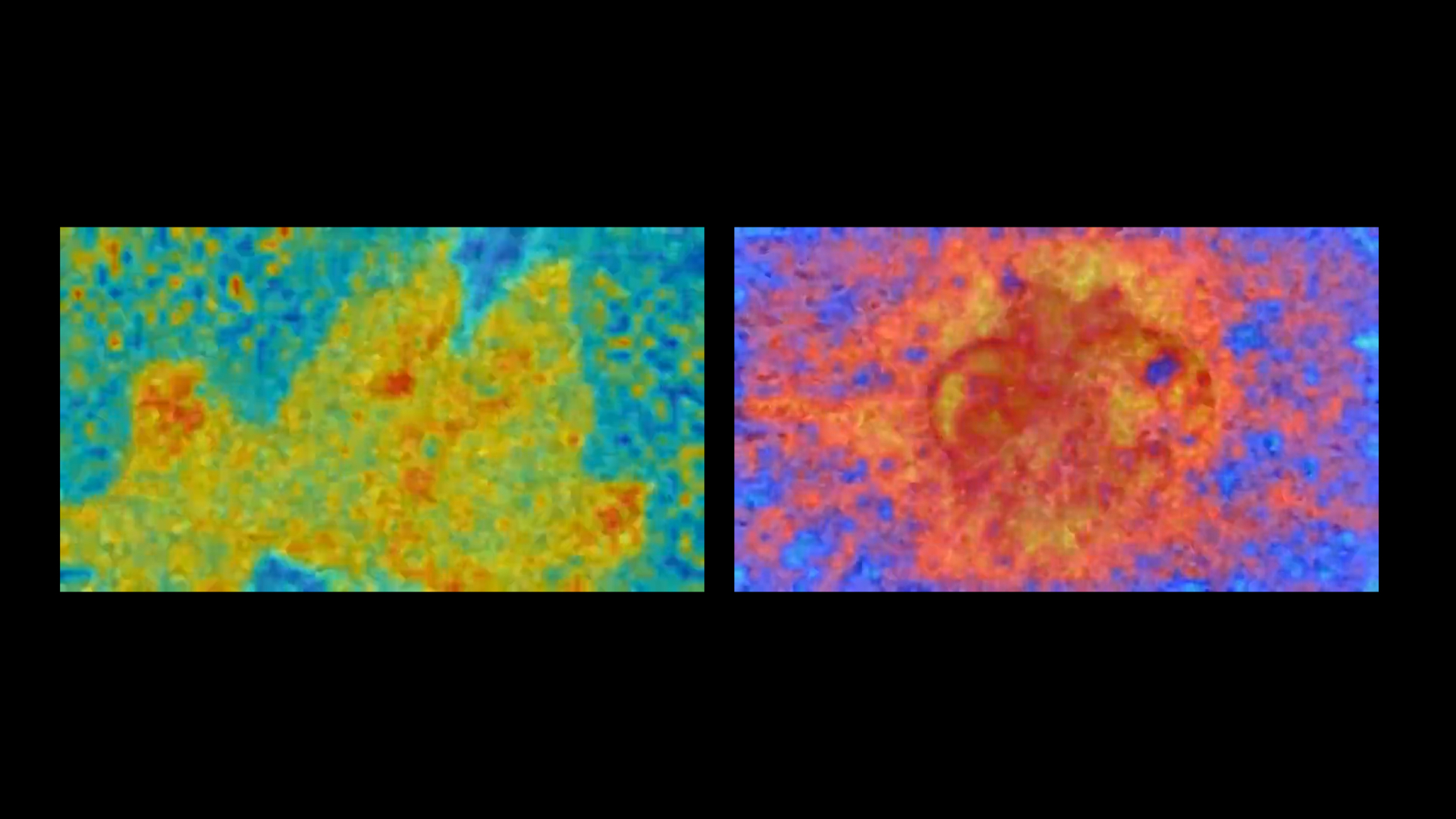}
  \caption{\textbf{Intercepting the signal.} Artistic study using attention maps from the internal diffusion process as creative material rather than the final video outputs. This diptych shows the structural similarities in the attention maps between an intimate kiss and a cosmic collision, revealing how the neural network sees \textit{impact} across different scales.}
\end{figure}

However, this ambition directly conflicts with the dominant interaction design paradigm for creative tasks involving AI \cite{torricelliRoleInterfaceDesign2024}. Most commercial creative AI tools rely entirely on prompt-based interfaces, a setup that keeps the internal mechanics of the model invisible and remote. Even open-source node-based interfaces like ComfyUI \cite{comfy-orgComfyUIMostPowerful2023}, which offer much greater visibility into the broader generation pipeline, abstract the core generative process into a singular ``sampler" node that reveals nothing more than a progress bar. Consequently, the true material of the tool remains inaccessible and opaque \cite{burrellHowMachineThinks2016}.  By reducing the artist's affordances strictly to textual inputs and pixel outputs, these systems inherently limit both the creative potential of the technology and the material intuition artists can develop while using it \cite{torricelliRoleInterfaceDesign2024, zamfirescu-pereiraWhyJohnnyCant2023}. 

This opacity is especially problematic in the realm of AI video generation. Video diffusion models have recently demonstrated remarkable fidelity in generating realistic moving images \cite{hoVideoDiffusionModels2022}, leading to a rapid proliferation of AI video across domains ranging from high-production films to social media \cite{zhangGenerativeAIFilm2025a, finneyAIScreenSector2025a, lyuPreliminaryExplorationYouTubers2024}. While industry and research communities are overwhelmingly focused on output quality performance \cite{ghildyalQualityPredictionAI2024}, such as the Frechet Video Distance (FVD) \cite{unterthinerAccurateGenerativeModels2019}, less focus has been given to understanding the internal mechanics of these models. The field of Explainable AI (XAI) \cite{arrietaExplainableArtificialIntelligence2019} seeks to address such gaps \cite{rudinInterpretableMachineLearning2021, abnarQuantifyingAttentionFlow2020a}, while the emerging subfield of Explainable AI for the Arts (XAIxArts) \cite{bryan-kinnsXAIxArtsManifestoExplainable2025} focuses specifically on the critical need for interpretability among artists engaging with generative systems.

One established interpretability method is the visualization of attention maps in transformer-based models, which can be rendered for human comprehension \cite{clarkWhatDoesBERT2019, cheferTransformerInterpretabilityAttention2021}. In text-to-image models like Stable Diffusion \cite{rombachHighResolutionImageSynthesis2022}, cross-attention maps highlight how specific text tokens in a prompt correspond to localized regions within a visual output \cite{tangWhatDAAMInterpreting2022a, helblingConceptAttentionDiffusionTransformers2025}. While these methods have been rigorously explored in image models, their application to video diffusion has only recently begun from a technical perspective \cite{liuUnderstandingAttentionMechanism2025, wenAnalysisAttentionVideo2025}, and remains profoundly underexplored from an artistic one.

To bridge this gap, this project introduces a tool for visualizing cross-attention in video diffusion transformers using the open-source video model Wan  \cite{wanteamWanOpenAdvanced2025}. The tool enables artists to systematically inspect attention behavior across heads (parallel attention sub-units), model layers (sequential transformer stages), and diffusion steps (iterations of gradual video refinement during generation). By making the internal process of the model systematically visible, we intend to help artists build a material intuition for the otherwise black-box diffusion process. We also demonstrate the creative value of this intervention through three artist studies that utilize the raw attention maps as artistic material. Altogether, bringing such transparency to the model offers a new vector for artistic experimentation, exposing how prompts shape generated videos and suggesting new strategies for creative intervention that go beyond the traditional outputs of AI systems.

\begin{figure}
    \centering
    \includegraphics[width=0.5\linewidth]{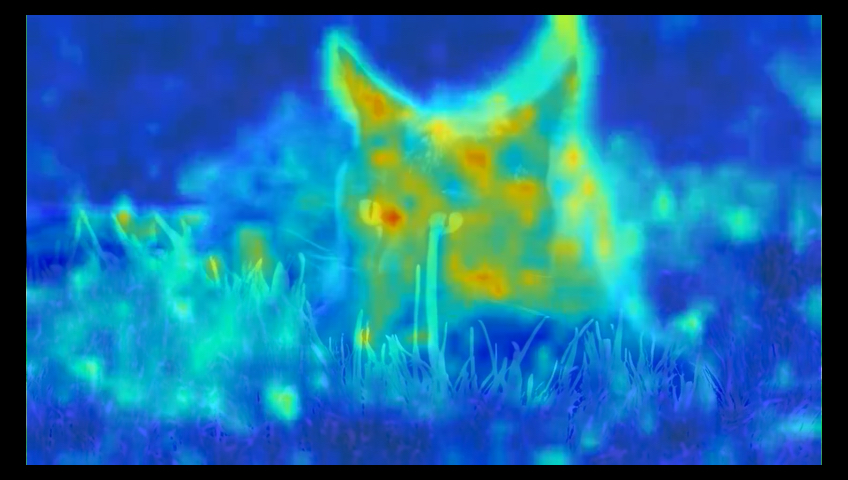}
    \caption{\textbf{Visualizing Attention}. A cross-attention map for the token ``cat" overlaid on the original generated video. The heatmap visualizes the spatial correspondence between text and image: warmer colors (yellow/orange) indicate regions where the token ``cat" has the strongest influence on the generated pixels, while cooler colors (blue) indicate areas of low association.}
    \label{fig:cat_single_attention}
\end{figure}

\section{Attention Maps: High Level Overview}

AI video models are increasingly built using a video diffusion transformer (DiT) architecture \cite{peeblesScalableDiffusionModels2023, openaiVideoGenerationModels2024, wanteamWanOpenAdvanced2025}. Rather than relying on traditional convolutional networks, these models flatten the spatio-temporal video latent space into a sequence of tokens, utilizing transformer blocks to iteratively denoise the representation \cite{vaswaniAttentionAllYou2017, peeblesScalableDiffusionModels2023, wanteamWanOpenAdvanced2025}. Within this architecture, attention maps serve as a central interpretability tool for understanding how these models operate across modalities, such as language and vision \cite{clarkWhatDoesBERT2019, cheferTransformerInterpretabilityAttention2021}.

The attention mechanism which underlies much of today's breakthroughs in AI is an elegant scaled dot-product:
\begin{equation}
  \text{Attention}(Q, K, V) = \text{softmax}\left(\frac{QK^T}{\sqrt{d_k}}\right)V
  \label{eq:attention}
\end{equation}
Cross-attention is the mechanism responsible for linking textual prompts to visual outputs, where these matrices take on distinct roles. The queries ($Q$) correspond to the embedded video representation (the spatio-temporal latents), while the keys ($K$) and values ($V$) correspond to the embedded prompt tokens, with $d_k$ representing the dimension of the key vectors. The cross-attention map itself is the weighting matrix produced by the softmax operation in Equation~\ref{eq:attention}. These attention weights indicate how much influence a given prompt token has on specific regions of the generated output. Crucially, this mechanism allows us to trace how specific words from the prompt map to regions of the visual outputs, effectively offering a window into the model's internal reasoning process.

In the broader field of XAI, attention maps have become a pivotal tool. This prominence is not because attention is the single most important element of the model's architecture, but rather because attention maps are one of the few natively human-interpretable elements of the computation. Because their dimensions map intuitively to the visual outputs, attention maps can be directly visualized as heatmaps over pixel space (ex. Fig.~\ref{fig:basic-attn-maps}). This unique property has driven significant research in the visual domain \cite{cheferTransformerInterpretabilityAttention2021, tangWhatDAAMInterpreting2022a, cheferAttendandExciteAttentionBasedSemantic2023, hertzPrompttoPromptImageEditing2022a, helblingConceptAttentionDiffusionTransformers2025}.

However, while these visualization techniques have been rigorously developed and utilized within the XAI research community to analyze and steer visual models, they remain under-explored from an artist's perspective. For practitioners utilizing generative AI as a medium, understanding and interacting with these inner mechanics offers a pathway to develop deeper creative intuition into the diffusion process and opens up unique artistic opportunities. By extracting and visualizing these maps for every prompt token across attention heads, attention layers, and diffusion steps, we can begin to better understand the generative decision-making structure and employ it directly as raw material for artistic exploration.

\section{Intercepting the Signal: Extraction and Visualization}

The focus of our study is the open-source video model Wan 2.1 \cite{wanteamWanOpenAdvanced2025}. We selected this model due to its widespread adoption within the research community \cite{chenSkyReelsV2InfinitelengthFilm2025, huangSelfForcingBridging2025, zhangVSAFasterVideo2025} and its elegant transformer architecture \cite{ wanteamWanOpenAdvanced2025}. As a latent diffusion model \cite{rombachHighResolutionImageSynthesis2022}, Wan operates within a compressed, lower-dimensional representation of video pixel space, condensing information both spatially and temporally. The 1.3 billion parameter variant utilizes 30 sequential transformer layer blocks, each housing a self-attention, cross-attention, and multi-layer perceptron layer; each cross-attention block features 12 distinct attention heads \cite{wanteamWanOpenAdvanced2025}. Consequently, for every single word-token in a prompt, the model calculates 360 distinct cross-attention maps per diffusion step. To access and visualize these internal calculations, we must intercept the transformer layer blocks actively during the generation process.

\begin{figure}[h]
    \centering
    \includegraphics[width=\linewidth]{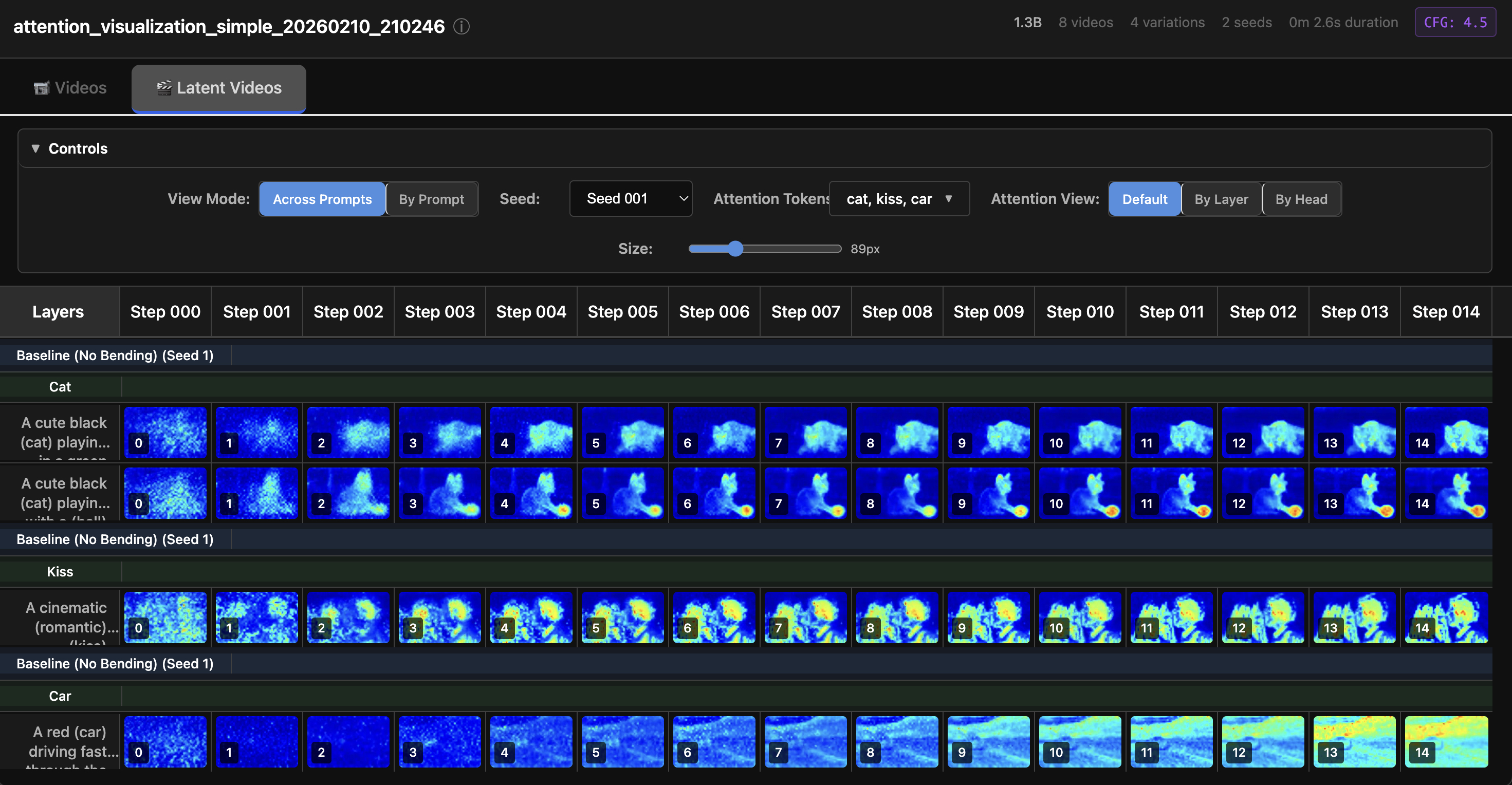}
    \caption{\textbf{Visualization Interface}. Web-based UI enabling both macro-level and granular exploration of cross-attention maps across diffusion steps, transformer layers, and individual attention heads. The default view shown here displays per-token attention maps averaged across each diffusion step: from left to right, the sequence reveals the development of attention from unfocused to concrete structure.}
    \label{fig:visualtion-interface}
\end{figure}

\subsection{Extraction and Decoding}
Our pipeline comprises two primary components: the \textit{extraction} of the raw attention data and its subsequent \textit{decoding} into human-readable visual outputs.

\textbf{Extraction:} We define a Python wrapper around the cross-attention layers of the Wan architecture. As the model generates a video, this wrapper intercepts each cross-attention computation and routes the data into local memory. To make the dimensionality of this data more legible, the final shape of these stored attention maps can be understood as an indexable five-dimensional tensor:
$$[ \text{DiffusionStep} \times \text{TransformerLayer} \times \text{AttnHead} \times \text{PromptToken} \times \text{FlatVideoLatent} ]$$
Because capturing this scale of data requires significant computational overhead, we conserve both VRAM and hard drive space by pre-selecting specific prompt tokens to track prior to generation.

\textbf{Decoding:} Once extracted, the raw attention maps must be decoded into a visual format. We first reshape the extracted maps from a flattened embedding into a 3D latent tensor. These tensors are then upscaled — spatially by a factor of 8 and temporally by a factor of 4 — to align with the dimensions of the final output video, rendering them as dynamic heatmaps. In these visual representations, brighter colors correspond to higher attention values, indicating areas of stronger influence, while muted-blue colors correspond to lower attention and less influence. We store these decoded videos for every tracked token across multiple levels of granularity: by diffusion step, by transformer layer, and by individual attention head. This systematic extraction yields a structured dataset containing potentially thousands of distinct attention videos for a single generated output. Next, we describe how we visualize these outputs in order to make sense of them.

\subsection{Interactive Exploration}
To make sense of this massive, multi-dimensional dataset and enable a rich understanding of the internal generative process, we developed a bespoke web-based visualization tool. As shown in Fig.~\ref{fig:visualtion-interface}, this interface allows for both a macroscopic view of per-token attention maps and a highly granular inspection, drilling all the way down to a specific attention head, at a specific transformer layer, during a precise diffusion step. 

In its default state, the tool displays the per-token attention maps across the chronological diffusion steps. These high-level maps are calculated by averaging the attention values across every head and layer for that specific step to provide legible overview. Alternatively, the \textit{Layer} view isolates the per-token attention maps averaged strictly across heads for a given block, while the \textit{Head} view exposes the raw, un-averaged computations made by every single individual attention head for a given token.

Ultimately, this tool is designed for interactive exploration. It allows artists to track the conceptual development of attention at a high level, while retaining the affordance to ask highly specific, low-level questions about how that ``meaning" was mathematically constructed step-by-step. By observing these emergent patterns firsthand, artist-practitioners can cultivate a hands-on intuition for the diffusion process. Furthermore, the interface acts as a material repository, providing easy access to the isolated attention map videos for use as raw material in downstream artistic works.

\section{Exploratory Probes}

To validate the effectiveness of visualizing cross-attention maps in video diffusion models, we conducted a series of exploratory probes. These tests were designed to assess increasing levels of complexity: single objects (a cat), multiple objects (a cat playing with a ball near the Eiffel Tower), objects in motion (a car on a highway), and abstract concepts or actions (a romantic kiss of people in love).

For these experiments, we utilized the Wan 1.3B parameter model, configured with 15 diffusion steps and a Classifier-Free Guidance (CFG) scale of 4.5 \cite{wanteamWanOpenAdvanced2025}. To increase robustness, we generated two seed variations per prompt (seeds 45 and 46), resulting in video outputs with a resolution of $848 \times 484$ pixels across 41 frames. By tracking the primary nouns and actions for each prompt, we generated a dataset covering 8 generation runs (4 prompts × 2 seeds), resulting in 93,840 unique attention video samples.


Below, we share high-level takeaways from our observations across these generations. Crucially, these observations are not offered as foundational proofs regarding the mathematical properties of the diffusion process. Rather, they provide a window into the intuition-building possibilities of investigating attention maps closely—a core tenet of XAIxArts practice.

\begin{table}[h]
\centering
\renewcommand{\arraystretch}{1.3} 
\caption{Experimental Dataset. A list of the exploratory probes used to validate the tool. Tracked tokens are highlighted in \textbf{bold} within the prompt text.}
\label{tab:probes}
\begin{tabularx}{\linewidth}{@{} l X c @{}} 
    \toprule
    \textbf{Probe Type} & \textbf{Prompt Text} \textit{(Tracked Tokens in Bold)} & \textbf{Figure} \\
    \midrule
    Single Object & A cute black \textbf{cat} playing in a green \textbf{field} & Fig. 2 \\
    Multiple Objects & A cute black \textbf{cat} playing with a \textbf{ball} in front of the \textbf{Eiffel Tower} & Fig. 4, 6 \\
    Motion & A red \textbf{car} driving fast through the rocky \textbf{mountains} & Fig. 5 \\
    Abstract Action & A cinematic \textbf{romantic} \textbf{kiss} between two people & Fig. 5 \\
    \bottomrule
    \end{tabularx}
\end{table}

\subsection{High Level Overview}
\begin{figure}[b]
    \centering
    \includegraphics[width=\linewidth]{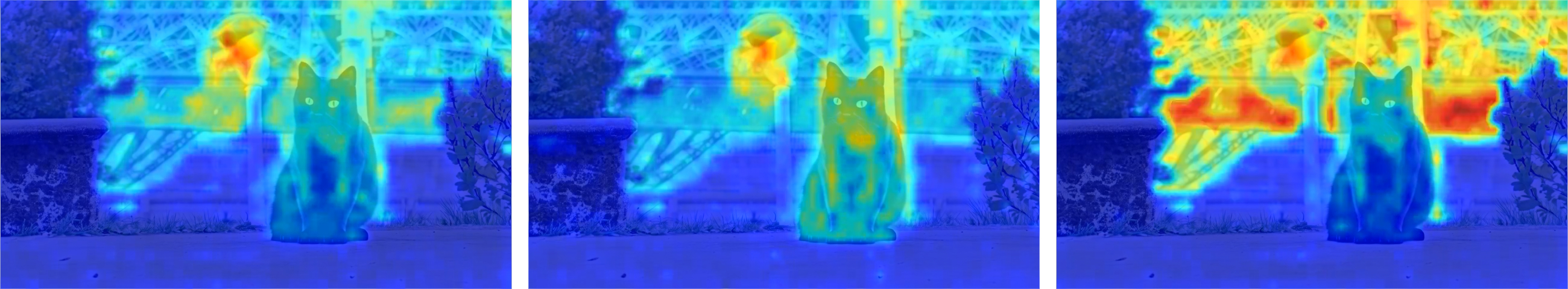}
    \caption{\textbf{Semantic separation and relational context.} Attention maps for ``cat" (left), ``ball" (center), and ``Eiffel Tower" (right). While the background architecture is clearly separated (right), the maps for the interacting subjects (cat and ball) exhibit significant overlap. This suggests the attention mechanism captures the relational context of the ``playing" action rather than strictly isolating object contours.}
    \label{fig:basic-attn-maps}
\end{figure}

\textbf{Single Object:} Using the prompt ``a cat," we confirmed that attention maps coherently align with object regions over time. The ``cat" token produced high attention values over the visual region occupied by the cat in the generated video, confirming the fundamental interpretability of Wan's attention maps (Fig.~\ref{fig:cat_single_attention}).

\textbf{Multiple Objects:} In a probe combining ``cat," ``soccer ball," and ``Eiffel Tower," we observed that each token’s attention localized to its corresponding object, demonstrating strong multi-object semantic separation (Fig.~\ref{fig:basic-attn-maps}). Notably, however, these segmentations were not always strictly bounded. For example, with the prompt ``cat playing with a ball," the attention regions for both ``cat" and ``ball" frequently overlapped, covering both objects simultaneously. This suggests that attention may not correspond strictly to the contours of isolated objects, but rather to the relational context in which they interact with other elements in the scene.

\begin{figure}[t]
    \centering
    \includegraphics[width=\linewidth]{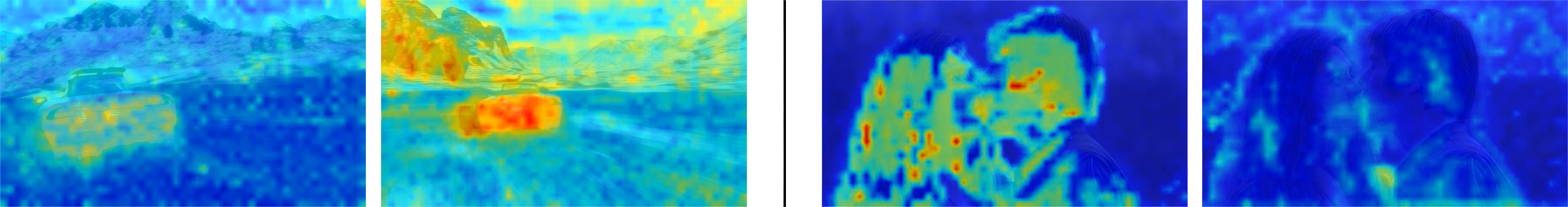}
    \caption{\textbf{Motion and Abstraction.} \textbf{Left:} Attention maps for the token ``car'' at two different timestamps. The model successfully tracks the object's movement across the frame despite changes in position. \textbf{Right:} A comparison of the concrete token ``kiss'' versus the abstract token ``romantic'' within the same frame. While ``kiss'' localizes to the physical interaction, ``romantic'' appears as a diffuse, scene-wide influence.}
    \label{fig:car-kiss}
\end{figure}

\textbf{Motion:} In Fig.~\ref{fig:car-kiss} (Left), we track the token ``car'' from the start and end of the generated video. The attention map consistently follows the object's trajectory, confirming that the model maintains semantic focus on moving targets. This sample also illustrates a phenomenon common to many of our generations: while the spatial focus remains consistent, the intensity of the attention map oscillates over time. This pulsing behavior may stem from the model's temporal compression (VAE) architecture, from artifacts inherent to the attention extraction process, or from other underlying dynamics of the attention mechanism. 

\textbf{Complex Actions and Abstract Concepts:} We further probed the model with the abstract concept of a ``romantic kiss'' (Fig.~\ref{fig:car-kiss}, Right). For the concrete token ``kiss,'' attention maps exhibited meaningful clustering around the subjects' lips and faces, although less spatially precise than those for physical objects in this study. In contrast, the map for the concept ``romantic'' was significantly more distributed. This observation highlights the diffuse manner in which abstract concepts and actions manifest within the attention mechanism, appearing as zones of interaction rather than rigid shapes.

\begin{figure}[h]
    \centering
    \includegraphics[width=0.55\linewidth]{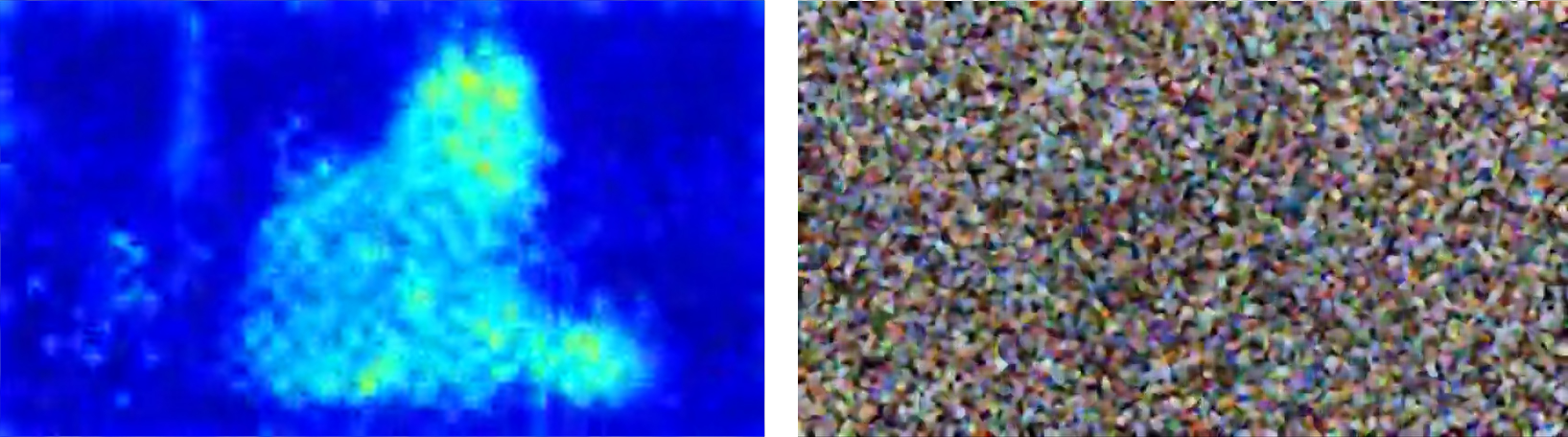}
    \caption{A comparison of the attention map (left) versus the decoded video content (right) at diffusion step 2 of 15. While the video output appears as unrecognizable noise, the attention map already displays strong compositional coherence, anticipating the final structure.}
    \label{fig:attention-vs-latent}
\end{figure}

\subsection{Process in Action}

\textbf{Across Steps:} visualizing the attention maps across the generation timeline reveals the diffusion process in action. As seen in Fig.~\ref{fig:visualtion-interface} from left to right, attention starts in a diffuse state but quickly conforms to the compositional elements of the final video. Remarkably, we can discern a high-level overview of the final composition in very early steps, even while the latent video itself visually resembles complete noise to the human eye (Fig.~\ref{fig:attention-vs-latent}). This suggests the degree to which the composition of a diffusion output is ``locked in" from the earliest moments of generation, reinforcing the intuition that diffusion develops from coarse global structure to fine local detail.

\begin{figure}[h]
    \centering
    \includegraphics[width=\linewidth]{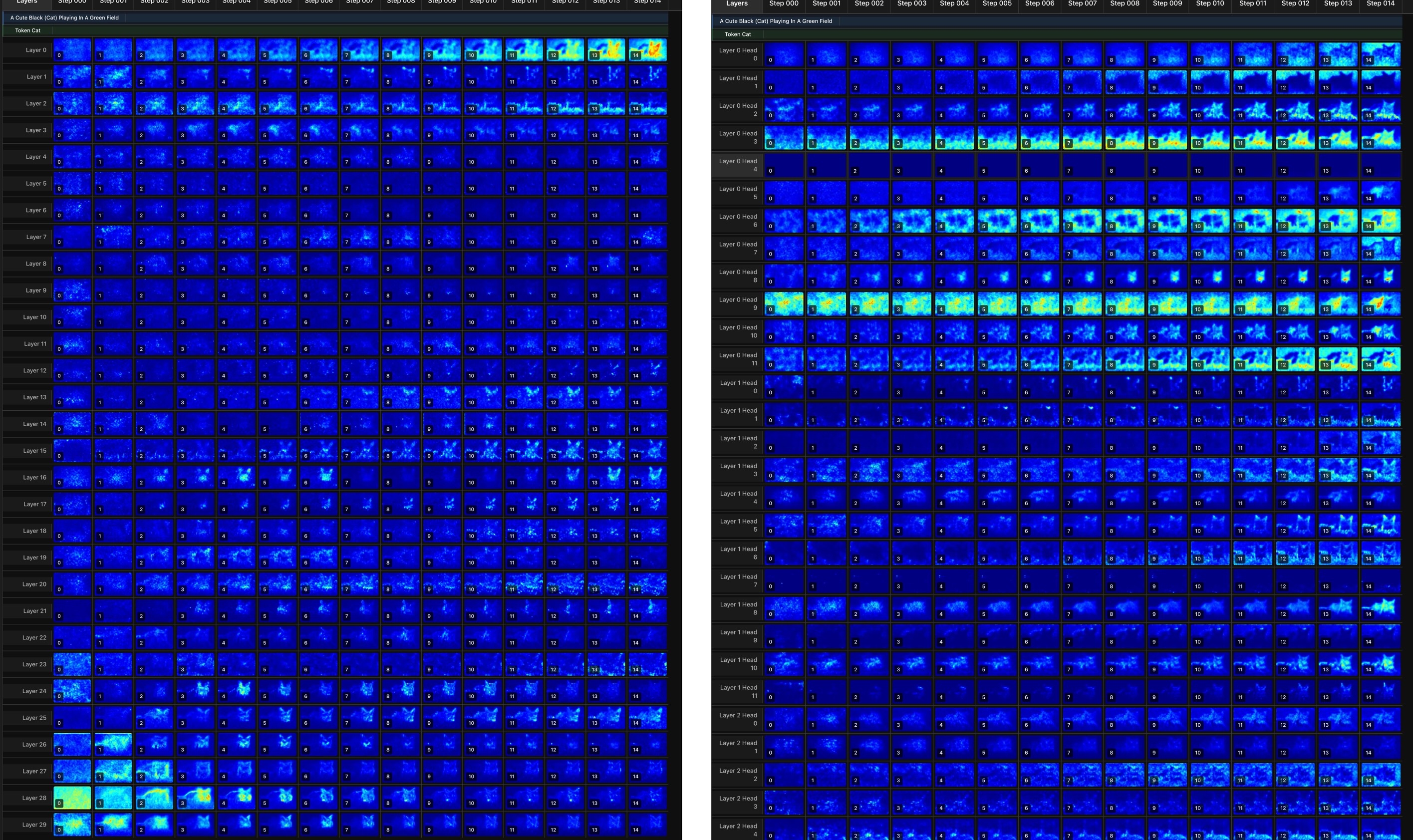}\caption{\textbf{Left:} Attention maps averaged by Layer, showing the evolution of attention across model depth. \textbf{Right:} Attention maps by Head, revealing the granular diversity of individual attention mechanisms for a single token.}
    \label{fig:layers-heads}
\end{figure}

\textbf{Across Layers:} The visualization in Fig.~\ref{fig:layers-heads} (Left) displays the dense landscape of information developed throughout the model's depth. While a definitive analysis of individual layers requires further study, this visual experiment offers high-level takeaways regarding the model's behavior. We visually observed consistent attention patterns in specific layers relative to the diffusion step. For instance, in our 15-step experiments, we consistently observed brighter regions of attention in the initial layers (0–3) during the later diffusion steps (13–15), whereas the final layers (27–30) were most active during the early diffusion steps (1–3). This observation suggests that while the specific information processed depends on the prompt, the structural role of these layers may be fairly consistent across diverse generation jobs.

\textbf{Across Heads:} Finally, drilling down to individual attention heads reveals the sheer density of computation involved in the diffusion process (Fig.~\ref{fig:layers-heads}, Right). Even without deducing the specific role of each head, we gain an intuition for the diversity of behavior within the model. For example, we observed that certain attention heads focus sharply on the primary object in the foreground, while others appear to contour to elements in the background. Similarly, some heads are significantly activated and dynamic, while many others remain muted or entirely flat throughout the generation, suggesting a sparsity in how the model utilizes its available attention capacity.

Collectively, these probes demonstrate how a deeper knowledge of the diffusion process can be developed through visualizing the cross-attention process. Next, we take this exploration a step further by demonstrating how this technical intuition can be operationalized to generate novel aesthetic forms.

\begin{figure}[b] 
    \centering 
    \includegraphics[width=\linewidth]{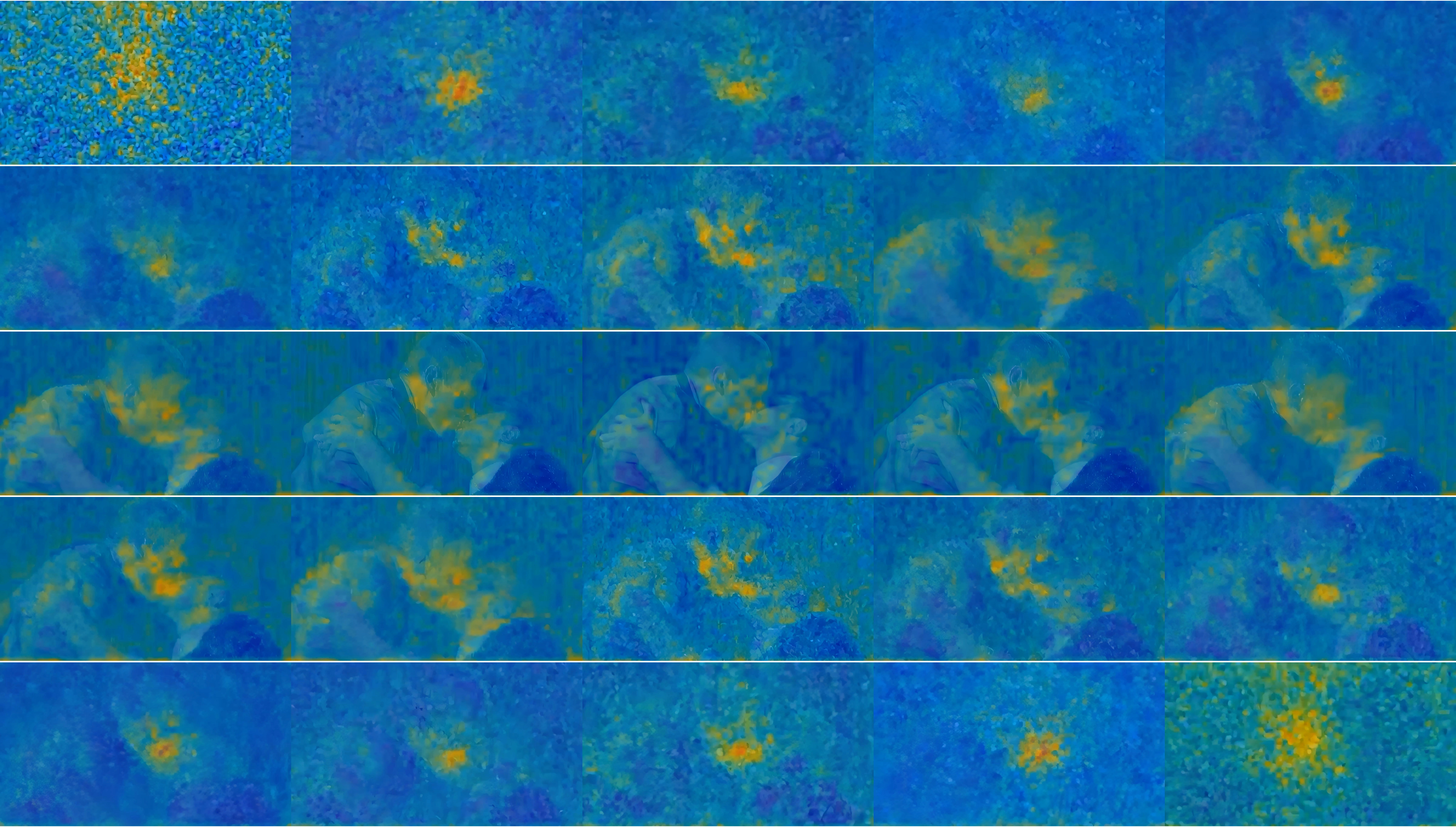} 
    \caption{\textit{Attention of a Kiss}: A video art study constructed from the attention maps of the token ``kiss" in the open source video model Wan. The sequence illustrates the transition from latent noise to semantic structure.}
    \label{fig:attention-of-a-kiss}
\end{figure}

\section[Artistic Explorations]{Artistic Explorations\footnote{Video documentation of the works described in this section is available at: \url{https://www.adamcole.studio/work/attention-of-a-kiss}.}}

One of the primary goals of this project was to create artistic outputs that both reveal the nature of the technology and disrupt its conventional aesthetics. A potent opportunity to achieve this lies in the repurposing of raw attention maps (usually discarded as ephemeral ``byproduct" data) as primary artistic material.

In a standard generation pipeline, these maps are calculated in a fraction of a second and immediately flushed from memory. By systematically capturing and archiving them, we are intentionally exploring a new visual language made possible only through this specific technical intervention. We present the following creative studies to illustrate the aesthetic and narrative potential of this approach.

\subsection{\textit{Attention of a Kiss}}

The video study \textit{Attention of a Kiss} (Fig.~\ref{fig:attention-of-a-kiss}) visualizes the evolving attention map of the specific token ``kiss" across the generation timeline. The video begins in total abstraction—a wash of latent noise—and gradually gains structure, paralleling both the diffusion process and the development of emotional intimacy. This metaphorical alignment between the model's construction of meaning and human interpretive processes suggests new narrative forms grounded specifically in AI diffusion mechanics.

\subsection{\textit{Multi-Head Attention}}

\begin{figure}[h] 
    \centering 
    \includegraphics[width=\linewidth]{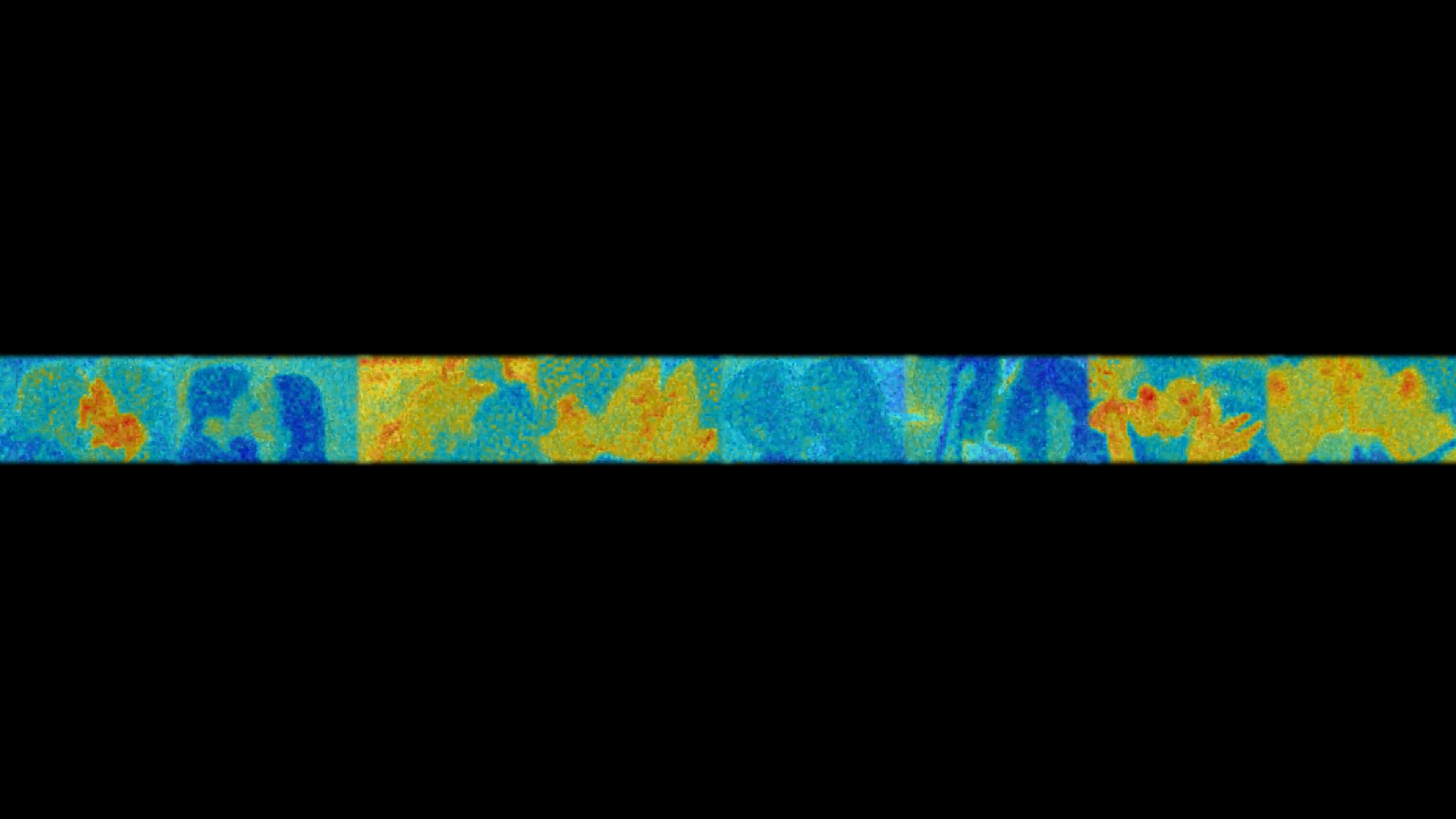} 
    \caption{\textit{Multi-Head Attention}: A visualization of the token ``cinematic" across eight different generations. The abstraction of the attention map obscures identity markers, reducing the scene to the gestural essence of the interaction.}
    \label{fig:multi-head-attention}
\end{figure}

In this study (Fig.~\ref{fig:multi-head-attention}), we extract the attention maps from the token ``cinematic" to visualize eight distinct kiss scenes as they develop over diffusion steps. While the output videos often echo familiar cinematic conventions of romance, the attention maps contain much more ambiguity. Conventional identity and gender markers are all but invisible in this abstracted view, especially at lower diffusion steps. Instead, we see the iconic image of the kiss reduced to a more mythical ideal. Repeated across scenes, the maps reveal both the flexibility of the pose and its underlying consistency, both physically and emotionally. Here, the ambiguity of the attention maps (as opposed to the specificity of the pixel-space outputs) offers a novel way to interpret the kiss beyond photorealistic representations.

\subsection{\textit{Diffusion Collision}}

\begin{figure}[h] 
    \centering 
    \includegraphics[width=\linewidth]{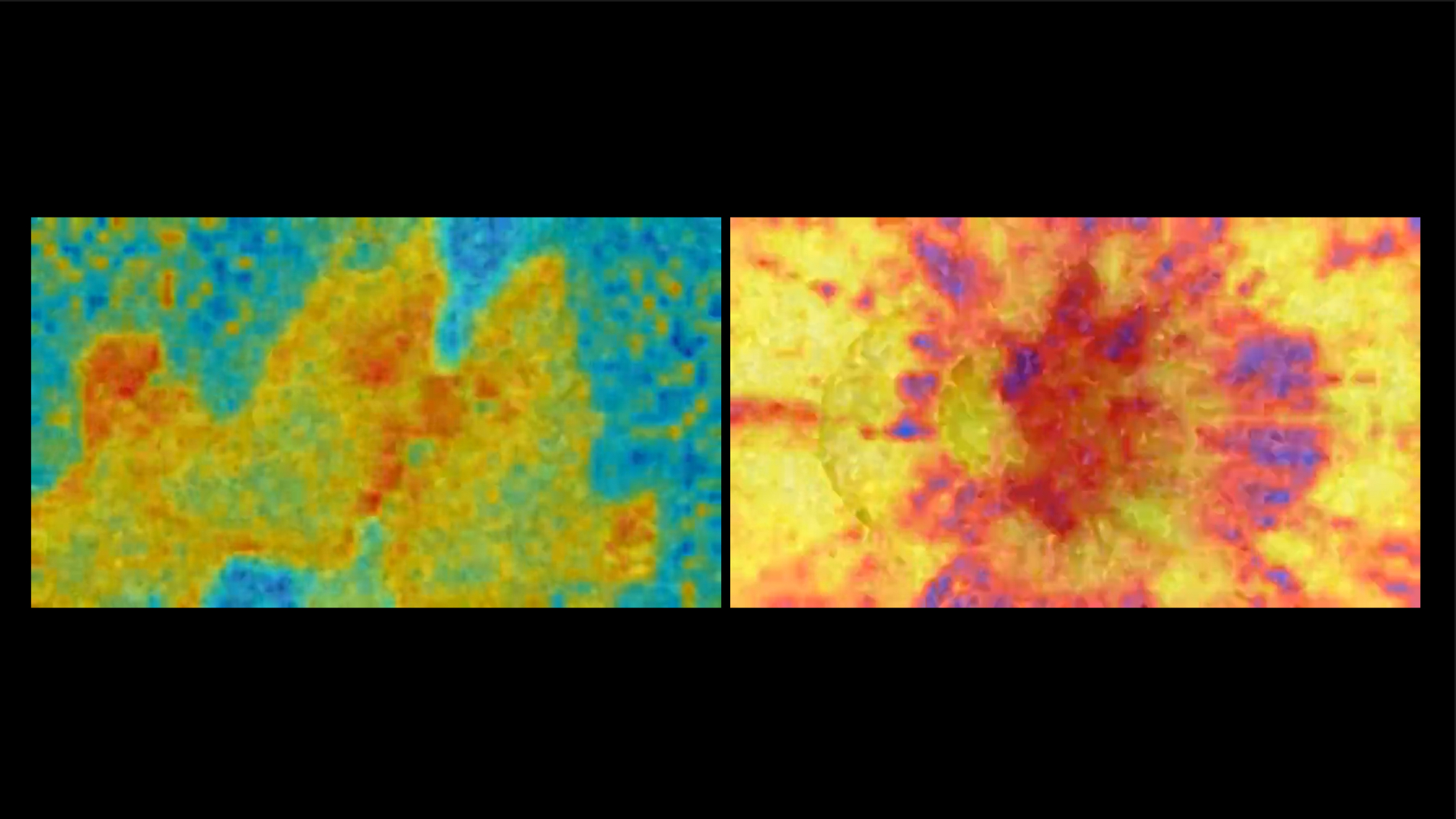} 
    \caption{\textit{Diffusion Collision}: A diptych comparing the attention dynamics of a human kiss (left) and a planetary collision (right). The study highlights the morphological similarities in how the model constructs ``impact" across vastly different semantic scales.} 
    \label{fig:diffusion-collision}
\end{figure}

Finally, in this diptych (Fig.~\ref{fig:diffusion-collision}), we pair two scenes of vastly different scales: on the left, the iconic kiss, and on the right, a cosmic collision between planets (with the attention colors inverted). At low diffusion steps across both videos, we see a commonality in the ambiguous forms developing in each scene and the rhythm of their impact. At higher diffusion steps, the distinction between the two scales and subject matters becomes clearer. Developing across the diffusion process, they create an interesting metaphor for the explosive collision of bodies, spanning from the personal to the galactic.

\section{Reflections}

\subsection{Usefulness of Attention Maps for Artists} 
Visualizing attention maps offers artists a valuable means of understanding how their textual prompts influence the visual outputs of generative video models. By revealing which regions of an image or video frame a given token attends to over time, attention maps allow artists to ``see what the model sees," making aspects of the model's internal generative process more tangible. This can help artists cultivate a more intuitive grasp of how prompts are parsed and interpreted.

As patterns emerge in how certain tokens correspond to specific visual traits, artists may begin to identify recurring motifs or dominant forms that shape the model’s output. For instance, an artist might investigate how the token “woman” translates into conventional visual markers of gender. This kind of feedback loop — between creative intent, expectation, and model behavior — can deepen the artist’s understanding of how language steers visual outcomes.

Beyond the functional benefit of prompting, this tool cultivates a form of \textit{material intimacy} with the generative system. Much like artists who develop a physical relationship to their medium in more traditional practices, like paint or clay, visualizing these internal calculations allows the digital artist to develop a tactile sense of the diffusion process. This material instinct is essential for artists seeking to push the tool beyond its normative defaults: bypassing the constraints of commercial user interfaces to engage directly with the fluid, probabilistic nature of the medium.

\subsection{Limitations and Future Work} 
While attention maps provide a useful lens into the internal mechanics of a generative model, they also have several limitations. In some test cases, the maps were noisy, inconsistent, or visually unintelligible, especially for tokens that were abstract or ambiguous. With longer prompts, multiple tokens often attend to overlapping regions, making it difficult to isolate the influence of each one. These issues highlight the importance of interpreting attention maps with care.

More broadly, attention maps reveal only a narrow aspect of the model’s behavior. While they capture token-to-region relationships, they do not account for broader elements such as compositional structure, representational logic, or temporal dynamics. As such, they should be treated as interpretive tools rather than comprehensive explanations of how generative models function. On a technical level, generating and analyzing attention maps remains resource-intensive, requiring significant GPU memory and producing large volumes of data that can be cumbersome to navigate.

It is also important to acknowledge the specific scope of this study. Our findings rely on observational evidence derived from a single model architecture (Wan 1.3B) and a limited set of prompt conditions. While these probes offer qualitative value for artistic intuition, they do not constitute a quantitative benchmark of the model's behavior. To encourage broader experimentation and future insights across different models and contexts, we are releasing our codebase as an open-source tool for the XAIxArts community to engage with.\footnote{Diffusion Latent Exploration: \url{https://github.com/adamcolestudio/diffusion_latent_exploration}}

\subsection{Conclusion: The Value of Understanding Your Tools} 
Just as early video artists built their own tools to understand and subvert the signal-based logic of analog video, artists today can gain creative leverage by exploring the inner mechanics of AI video models. Attention maps offer one such entry point — revealing how specific language tokens modulate the generation process over space and time. However, this is only the beginning.

A deeper understanding of a model's internal architecture can inspire a new generation of media practices that operate not just on the outputs of models but within their internal logic. This opens the door to network bending \cite{broadNetworkBendingExpressive2021}, where the structure and flow of computation within generative models are re-imagined as artistic parameters.

By treating the neural network itself as a malleable medium, artists can step beyond prompt engineering to creatively intervene in the generation process, producing outputs that go beyond the intended domain of the model. These explorations extend the lineage of experimental video art into the realm of generative AI, where the artwork emerges not only from what is seen, but from how the network sees.

%
%
\bibliographystyle{spmpsci}
\bibliography{PhD-refs}

\end{document}